# The Weather Paradox: Why Precipitation Fails to Predict Traffic Accident Severity in Large-Scale US Data

Yann Bellec[3], Rohan KAMAN[1], Siwen CUI[1], Aarav AGRAWAL[2], Calvin CHEN[1]

[1] Department of Computer Science, University of California San Diego, 9500 Gilman Dr, La Jolla, CA 92093
[2] Department of Data Science, University of California San Diego, 9500 Gilman Dr, La Jolla, CA 92093
[3] Department of Neuroscience, University of California San Diego, 9500 Gilman Dr, La Jolla, CA 92093
* Corresponding author, e-mail: rkaman@ucsd.edu, s4cui@ucsd.edu, cac028@ucsd.edu, a7agrawal@ucsd.edu, ybellec@ucsd.edu

**Abstract**
This study investigates the predictive capacity of environmental, temporal, and spatial factors on traffic accident severity in the United States. Using a dataset of 500,000 U.S. traffic accidents spanning 2016–2023, we trained an XGBoost classifier optimized through Randomized Search Cross-Validation and adjusted for class imbalance via class weighting. The final model achieves an overall accuracy of 78%, with strong performance on the majority class (Severity 2), attaining 87% precision and recall. Feature importance analysis reveals that time of day, geographic location, and weather-related variables, including visibility, temperature, and wind speed, rank among the strongest predictors of accident severity. However, contrary to initial hypotheses, precipitation and visibility demonstrated limited predictive power, potentially reflecting behavioral adaptation by drivers under overtly hazardous conditions. The dataset's predominance of mid-level severity accidents constrains the model's capacity to learn meaningful patterns for extreme cases, highlighting the need for alternative sampling strategies, enhanced feature engineering, and integration of external datasets. These findings contribute to evidence-based traffic management and suggest future directions for severity prediction research.

## 1.1 Context and Significance

Traffic accidents represent a significant public health and safety concern worldwide, causing approximately 1.35 million deaths globally in 2016 and constituting the leading cause of death among individuals aged 5 to 29 years [1]. Beyond direct physical and financial harm to those involved, traffic accidents substantially disrupt transportation networks and extend commute times, imposing broader societal costs. Understanding the conditions that contribute to accident occurrence and severity is therefore essential for improving public welfare and minimizing disruptions to daily life. Given these implications for public safety and transportation management, traffic accident severity prediction has attracted extensive research attention [11, 13].

## 1.2 Related Work

Recent scholarship has emphasized the importance of geographic context in accident prediction. Zhi (2025), employing interpretable machine learning models focused on California, demonstrated that risk factors vary significantly across locations and that the significance of specific conditions depends heavily on geographic context [2]. This finding underscores the heterogeneity of risk factors and the limitations of generalized assumptions in safety modeling.

Methodological advances have also addressed the challenge of imbalanced datasets. Research published in Scientific Reports developed neural network methods for handling the inherent imbalance in accident data, where severe accidents occur far less frequently than minor incidents [3]. These studies revealed that weather and visibility are influential factors in severe accidents while highlighting the critical need for awareness of data distribution when modeling traffic outcomes.

Complementary research conducted in Jordan analyzed 177,378 accidents between 2016 and 2021 using Geographic Information Systems, kernel density estimation, and Random Forest classification with Bayesian hyperparameter optimization, achieving 90% accuracy [4]. Notably, this study identified temporal factors as the third most important variable defining severity, with accident spikes occurring on Mondays and Fridays, during June and August, and particularly during the 2–4 PM time window. The authors attributed these patterns to reduced alertness, fatigue, and increased traffic volume. Importantly, the study revealed a counterintuitive negative correlation between poor weather conditions and accident severity, explained by increased driver caution and adaptive behavior under perceived risk.

Additional research from Iran employing the Analytic Network Process method prioritized environmental risk factors, identifying road slipperiness as the most critical environmental sub-factor, followed by road surface conditions and traffic lane type [5].

**1.3 Research Question and Hypothesis**

This study addresses the following research question: Can the severity of traffic accidents, operationalized as the level of disruption or impact on traffic flow, be predicted using environmental, temporal, and spatial factors such as weather conditions, visibility, precipitation, and time of day? Furthermore, which factors most strongly influence the likelihood of high-severity incidents?

We hypothesize that environmental, spatial, and temporal conditions, particularly precipitation, visibility, and time of day, are significant predictors of traffic accident severity [7, 17, 18]. Poor visibility and heavy precipitation are expected to increase severity by reducing drivers' ability to perceive hazards and lengthening reaction times, thereby making collisions more difficult to avoid and resulting in more extensive traffic disruption [19, 20].

Time of day is also expected to influence accident severity. Incidents occurring during peak commuting hours are hypothesized to have greater impact due to higher traffic density, limited maneuvering space, and slower emergency response times [21, 22]. Conversely, accidents during nighttime or low-traffic periods may involve higher driving speeds but affect fewer vehicles overall, potentially producing shorter but more severe disruptions depending on road type and visibility [23, 26].

We acknowledge the possibility of behavioral adaptation, wherein drivers mitigate severity through heightened caution under adverse conditions. Previous studies have reported mixed findings regarding driver adaptation in poor weather, with some evidence suggesting that drivers become more cautious, reducing crash likelihood but not necessarily mitigating severity once accidents occur [31, 33]. By examining these relationships across varied conditions, we aim to test whether observed patterns hold consistently and to identify the combination of factors most strongly predictive of high-severity outcomes [35, 36].

**2.1 Dataset Description**

The dataset employed in this study is the US Accidents (2016–2023) compilation by Moosavi, hosted on Kaggle [6]. The complete dataset contains over 7.7 million traffic accident records with 46 variables, collected from the U.S. Department of Transportation, state transportation agencies, and traffic monitoring infrastructure. To ensure computational efficiency, we utilized a representative sample of 500,000 entries provided by the dataset curators, retaining all 46 original variables.

Each observation represents a verified traffic incident with detailed environmental, temporal, and spatial information. Key variables relevant to our research question include: Severity (1, 4), a categorical indicator of accident impact level where 1 represents least impact on traffic and 4 represents greatest impact; Start_Time and End_Time for capturing temporal patterns; Start_Lat and Start_Lng for geographic coordinates; and environmental variables including Temperature (°F), Visibility (mi), Precipitation (in), and Wind_Speed (mph). Additional contextual indicators include Sunrise_Sunset and Wind_Direction. The dataset's national coverage and substantial sample size enable analysis of long-term temporal patterns, though the sampling approach may amplify the effects of inherent data missingness.

**2.2 Variable Operationalization**

Severity quantifies accident impact on traffic on a scale of 1 to 4, measured by traffic delay duration [41]. Geographic variables include street, city, county, state, ZIP code, and time zone, providing spatial context for each accident site [42, 44]. Weather_Timestamp indicates the time of weather observation in local time.

Temperature and Wind_Chill are measured in Fahrenheit, with values below 40°F indicating cold conditions and values above 90°F indicating hot conditions [45, 46]. Humidity, expressed as a percentage, represents moisture content relative to maximum capacity at current temperature, with values below 30% indicating low humidity and above 80% indicating high humidity [47, 48]. Pressure measures atmospheric pressure in inches of mercury, with values above 29.8 generally indicating fairer weather and stable air at sea level.

Visibility, measured in miles, represents horizontal distance at which objects can be discerned by the average human eye [49]. Wind_Direction uses standard abbreviations (N, S, E, W), while Wind_Speed in miles per hour classifies conditions from calm (~0 mph) through gales (40, 50 mph) to storm conditions (>50 mph). Precipitation quantifies rainfall in inches, with values below 0.1 inches classified as light rain and above 0.3 inches as heavy rain. Time of day is categorized through Sunrise_Sunset based on sunrise and sunset times, with Civil_Twilight, Nautical_Twilight, and Astronomical_Twilight providing additional categorizations based on solar position.

**2.3 Data Quality Assessment and Preprocessing**

Initial assessment revealed substantial class imbalance in severity distribution. Severity level 2 dominates the dataset (397,538 entries, 79.62%), while levels 4 (13,040 entries, 2.61%) and 1 (4,260 entries, 0.85%) are substantially underrepresented. This imbalance was addressed through class weighting during model training, enabling retention of sufficient data for statistical power [50, 52].

Examination of continuous variables revealed implausible outliers suggesting data entry errors, including temperatures of 207°F and wind speeds of 822.8 mph [53]. While extreme outliers are readily identifiable and removable, the presence of such values raises concerns about systematic recording errors [54, 55]. To mitigate this potential bias, we validated data against realistic physical limits using IQR-based outlier filtering [56, 58].

Substantial missing data was identified for Precipitation (142,563 missing values) and Wind_Speed (36,958 missing values) [59, 60]. Given our hypotheses regarding weather conditions as predictors, appropriate handling of these missing entries was essential [61]. Additionally, the dataset's aggregation from multiple APIs and agencies introduces potential inconsistencies in variable standards across sources [62, 63], which may introduce geographic biases in severity reporting.

Outliers were examined against established thresholds: Temperature < −50°F (Alaska winter minima), Visibility > 90 mi (EPA national park visibility ranges), Wind speed > 130 mph (Category 4 hurricane threshold per Saffir–Simpson scale), and Precipitation > 4 in/observation (based on World Meteorological Organization classifications). Cross-referencing with historical weather records via GPS coordinates confirmed that several extreme values, including a −77.8°F reading in New York City, visibility measurements ≥100 mi, and wind speeds of 131–142 mph without corresponding storm activity were recording errors requiring removal.

**2.4 Statistical Analysis and Modeling Approach**

Exploratory data analysis comprised univariate analysis (examining distributions via histograms and boxplots to assess class imbalance and detect outliers [64]), bivariate analysis (exploring correlations between weather features and accident severity [65, 66]), and temporal/categorical analysis (investigating time-based patterns and categorical features [67, 68]).

For predictive modeling, we implemented a Weighted XGBoost classifier to address class imbalance. Class weights were applied to prioritize detection of rare, high-severity accidents over accuracy maximization on the majority class. Hyperparameter optimization was conducted via Randomized Search Cross-Validation.

**3.1 Univariate Distributions**

Severity distribution analysis confirmed substantial class imbalance: Severity 2 comprised 79.62% of records (n = 397,538), Severity 3 comprised 16.92% (n = 84,477), Severity 4 comprised 2.61% (n = 13,040), and Severity 1 comprised 0.85% (n = 4,260) of the 499,315 total accidents.

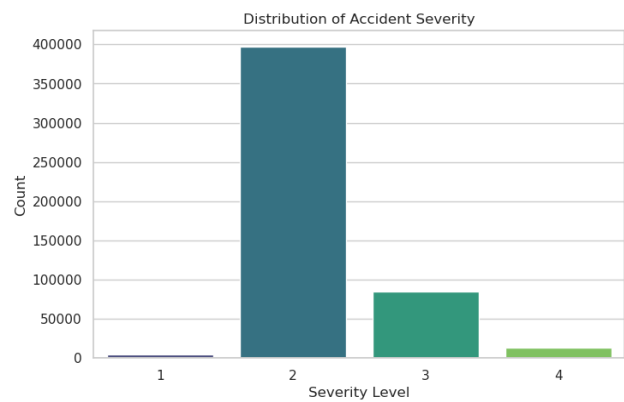

**Fig. 1.** Distribution of accident severity levels (1–4) in the dataset. Severity 2 dominates with 79.62% of records (n = 397,538), while Severity 1 (0.85%) and Severity 4 (2.61%) are substantially underrepresented.

Analysis of key numerical features revealed characteristic distributional patterns. Temperature exhibited moderate left skew with a mean of 61.7°F. Visibility, wind speed, and precipitation demonstrated substantial right skew, consistent with physical constraints preventing negative values while permitting extreme positive values. Median values (used given skewness) indicated visibility centered at 10 mi, wind speed at 7 mph, and precipitation near 0 in. The high skewness coefficients for visibility (2.72) and precipitation (91.76) suggested presence of unrealistic extreme values, which subsequent outlier analysis confirmed.

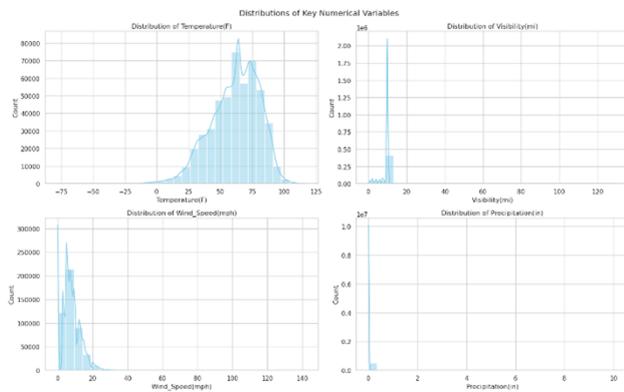

**Fig. 2.** Histograms of key meteorological variables. Temperature exhibits moderate left skew; visibility, wind speed, and precipitation show pronounced right skew consistent with physical constraints.

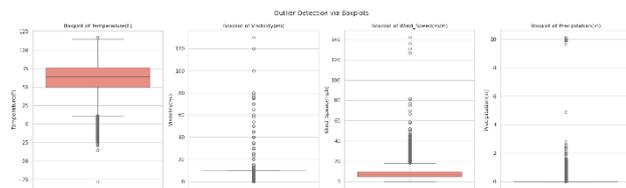

**Fig. 3.** Boxplot-based outlier detection for meteorological variables. Outliers were identified using IQR criteria and cross-referenced with historical weather records.

### 3.2 Temporal and Categorical Patterns

Temporal analysis extracted month, hour, and weekday from event timestamps. Severity displayed clear seasonal patterns, gradually increasing through spring, peaking during mid-summer (June–August), and declining toward winter, suggesting that warmer months may be associated with behavioral patterns resulting in more severe crashes.

Daily patterns revealed more pronounced differences. Severity reached its nadir during early morning hours (2–5 AM), rose through late morning, remained elevated during afternoon hours, and peaked in the evening around 7–8 PM following typical commuting hours. This spike likely reflects the combined effects of traffic density, post-workday fatigue, and reduced visibility. In contrast, weekday variations were minimal, suggesting limited influence of day-of-week on severity relative to time-of-day and seasonal effects.

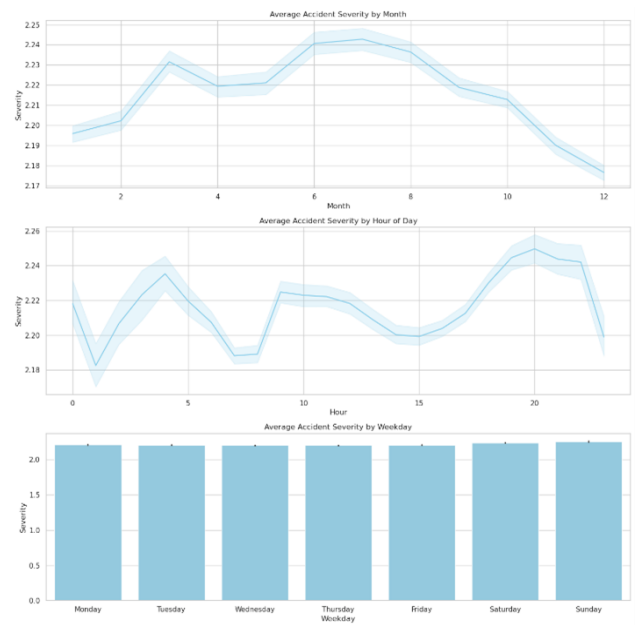

**Fig. 4.** Temporal variation in mean accident severity. Top: seasonal pattern with summer peak (June–August). Middle: daily pattern with evening peak (~7–8 PM). Bottom: minimal weekday variation.

Categorical analysis of Wind Direction and Sunrise/Sunset revealed modest variation in severity across wind directions, though CALM conditions showed highest average severity, suggesting severe accidents are not uniquely linked to high-wind scenarios. The Sunrise/Sunset comparison showed approximately equal severity between daytime and nighttime accidents, contradicting the common assumption that nighttime accidents are inherently more severe due to visibility constraints.

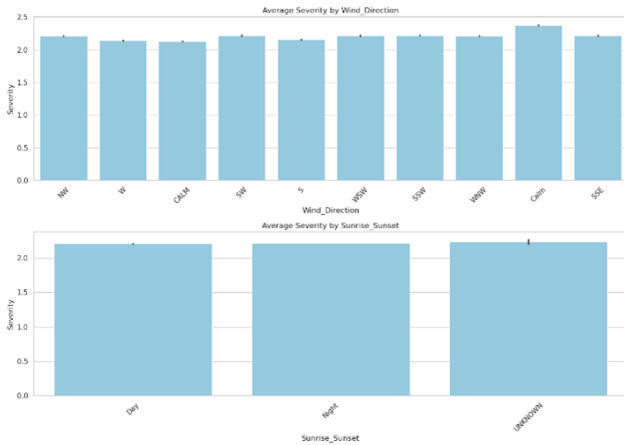

**Fig. 5.** Mean accident severity by wind direction (top) and daylight condition (bottom). CALM conditions show highest severity; daytime and nighttime accidents exhibit comparable severity levels.

### 3.3 Bivariate Relationships

Analysis of relationships between meteorological factors and accident severity revealed minimal association. Boxplots demonstrated similar distributions across all four severity strata for temperature (wide dispersion from negative values to >100°F with similar medians), visibility (concentrated around 10 mi with values below 1 mi rare), wind speed (predominantly weak winds between 0–12 mph), and precipitation (highly asymmetric distribution centered near zero).

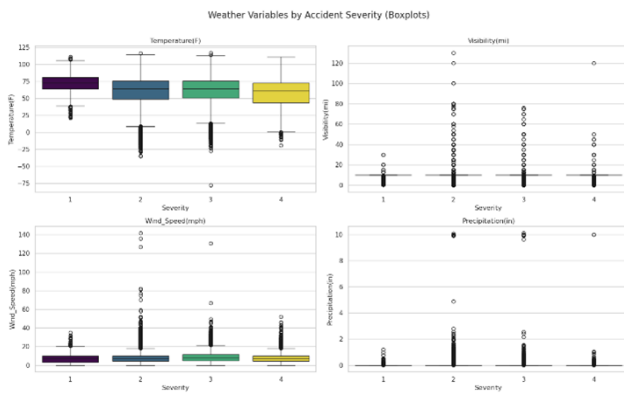

**Fig. 6.** Boxplots of meteorological variables stratified by accident severity. Similar distributions across all severity levels suggest limited discriminative power of weather features.

Density kernel estimates confirmed these patterns: temperature distributions showed similar shapes across severity levels with concentrated central masses and extended tails; visibility densities were primarily located at high values without apparent relationship to severity; wind speed observations formed narrow cores with extended tails unassociated with severity; and precipitation distributions were highly localized near zero without illuminating severity relationships.

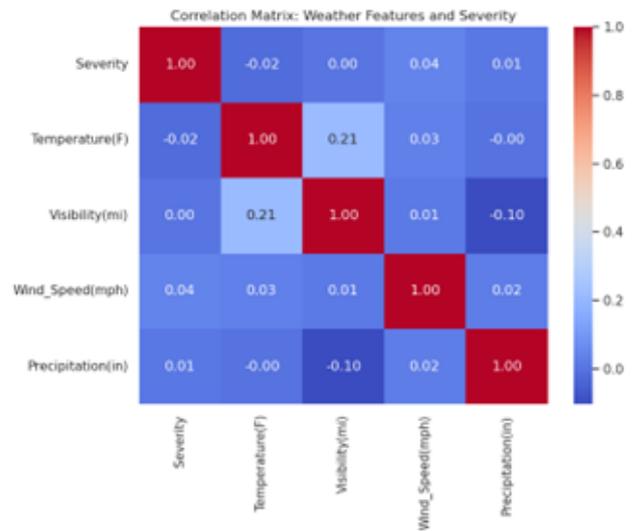

**Fig. 7.** Pearson correlation matrix between severity and meteorological variables. Near-zero correlations confirm absence of linear relationships between weather conditions and accident severity.

Correlation analysis yielded coefficients near zero between severity and all meteorological variables, confirming absence of exploitable linear relationships. Internal meteorological correlations were also weak, except for a moderate temperature-visibility correlation that showed no impact on severity. These bivariate results demonstrate that the measured meteorological variables exhibit no exploitable relationship with accident severity in this dataset, with distributions showing stable patterns across severity modalities.

### 3.4 XGBoost Model Performance

The Weighted XGBoost classifier achieved overall accuracy of 78.2% with distinct performance characteristics across severity levels. For Severity 2 (majority class), the model demonstrated high reliability with precision and recall of 0.87, indicating effective learning of standard traffic incident patterns. For Severity 3, the model achieved recall of 0.49, successfully identifying nearly half of serious accidents and confirming that specific environmental and temporal features serve as meaningful predictors of increased severity.

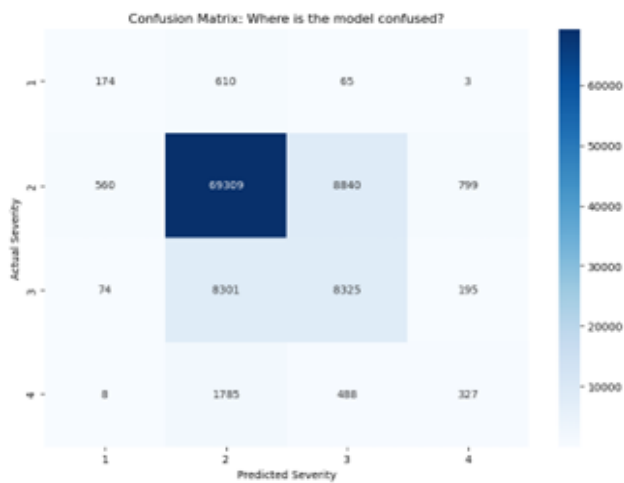

**Fig. 8.** Confusion matrix for the Weighted XGBoost classifier. Strong diagonal dominance for Severity 2 (precision/recall = 0.87); substantial misclassification of Severity 4 cases as Severity 2 reflects class imbalance effects.

For Severity 4 (fatal/critical accidents), model recall was 0.13, lower sensitivity but still providing non-zero signal for the most critical events. Confusion matrix analysis revealed that 1,785 of 2,608 Severity 4 accidents were misclassified as Severity 2. This reduced performance at the extreme suggests that fatal accidents likely depend on variables absent from this dataset (e.g., driver behavior, vehicle speed, intoxication) rather than environmental conditions alone.

Feature importance analysis identified the five strongest predictors: start site latitude, temperature, start site longitude, wind speed, and hour of day. These results largely support our hypothesis regarding the predictive value of temporal and spatial factors while revealing that geographic coordinates contribute substantially more predictive information than anticipated.

### 4.1 Interpretation of Findings

This study demonstrates that traffic accident severity can be predicted from environmental, spatial, and temporal features with moderate accuracy using machine learning approaches. The model's strong performance on Severity 2 accidents (precision and recall of 0.87) confirms its capacity to accurately classify the majority of traffic incidents. The identification of latitude, longitude, temperature, wind speed, and hour as principal predictors largely supports our hypothesis regarding the importance of temporal and weather factors while highlighting the substantial contribution of geographic location.

Notably, precipitation and visibility, variables more directly reflecting hazardous driving conditions, did not emerge as significant predictors. While initially counterintuitive, this finding aligns with previous research demonstrating behavioral adaptation: drivers may exercise heightened caution under overtly poor conditions, reducing crash severity even when accident occurrence is unaffected. In contrast, temperature and wind speed, which are less commonly interpreted as warning signals, may serve as proxies for specific environmental conditions influencing severity without triggering compensatory vigilance.

The strong predictive contribution of geographic coordinates reinforces Zhi's (2025) findings on the centrality of location in accident risk assessment. The prominence of temporal features corroborates results from the Jordan road-accidents study, which similarly identified time-of-day as a key severity determinant. The evening severity peak around 7–8 PM likely reflects compounded effects of traffic congestion, driver fatigue, and transitional lighting conditions.

### 4.2 Limitations

The severe class imbalance in our dataset constitutes the primary methodological limitation. With Severity 2 comprising 79.62% of observations and Severity 4 only 2.61% (n = 2,608 in test set), the model lacks sufficient representation of extreme cases to learn distinguishing patterns effectively. Despite class weighting, performance remains constrained at the extremes, and features identified as important may not generalize to trivial or critical accidents.

The natural rarity of critical accidents combined with likely underreporting of trivial incidents explains this distributional skew. The bivariate analysis further revealed that weather variables in this dataset cannot adequately capture the context needed for severity prediction: precipitation and wind speed are dominated by near-zero values, preventing proper study of extreme conditions, while visibility is almost always favorable.

These findings are consistent with broader literature indicating that weather variables explain only approximately 6.9% of variance in traffic fatalities, with 86.6% of accidents occurring under normal conditions without clear weather-severity linkage [5]. The effects of weather are heterogeneous, sometimes paradoxical, and strongly dependent on accident type, driver characteristics, and road network configuration.

**4.3 Ethical Considerations**

Several ethical dimensions warrant consideration. First, the data collection process may lack transparency regarding informed consent from individuals involved in documented accidents. Second, geographic bias may arise from uneven distribution of traffic monitoring infrastructure, potentially underrepresenting remote or economically disadvantaged areas. Third, using traffic impact duration as a severity proxy may introduce systematic biases not captured by our modeling approach.

The publicly available dataset contains no direct personally identifiable information, with all location, time, and environmental data pre-anonymized. Nevertheless, model outputs should be protected and restricted to academic use to prevent misapplication in legal or financial contexts, such as biased insurance pricing or claim evaluation.

**4.4 Future Directions**

Addressing the class imbalance limitation will require targeted data acquisition strategies focusing on extreme-severity cases, advanced resampling techniques, and integration of behavioral variables currently beyond passive sensor network capabilities. Future research should investigate whether specific patterns in geographic location, temperature, wind speed, and time correlate with more severe accidents across balanced severity distributions.

From an applied perspective, these results support development of real-time predictive systems incorporating spatiotemporal risk patterns to enable dynamic interventions, from adaptive signaling to targeted driver alerts, deployed precisely when and where severity risk peaks. Integration of real-time weather data into traffic management systems, combined with engineering measures such as dynamic signage and context-dependent speed limits, represents a promising direction for severity reduction.

**5 Conclusion**

This study demonstrates that traffic accident severity in the United States can be moderately predicted from environmental, spatial, and temporal features using machine learning approaches. Our XGBoost classifier, optimized for class imbalance, achieves 78% accuracy and identifies geographic coordinates, temperature, wind speed, and time of day as dominant predictors while, counterintuitively, visibility and precipitation exert minimal predictive influence. This pattern aligns with emerging evidence that drivers adapt behavior under overtly hazardous conditions, whereas subtler environmental cues escape compensatory vigilance.

The severe underrepresentation of extreme-severity events constrains model generalization to the most consequential accidents, underscoring a fundamental limitation inherent to observational traffic data. Addressing this gap will require targeted data acquisition, advanced resampling methodologies, and integration of behavioral variables currently beyond passive monitoring capabilities.

Beyond methodological refinement, these results carry immediate implications for evidence-based traffic management. Real-time predictive systems informed by spatiotemporal risk patterns could enable dynamic interventions precisely when and where severity risk peaks. As urban mobility systems grow increasingly complex, such data-driven approaches offer a scalable pathway toward measurably safer roads.